\documentclass[conference]{IEEEtran}
\IEEEoverridecommandlockouts
\usepackage{cite}
\usepackage{amsmath,amssymb,amsfonts}
\usepackage{algorithmic}
\usepackage{graphicx}
\usepackage{textcomp}
\usepackage{xcolor}

\usepackage{booktabs}
\usepackage{multirow}
\usepackage{tabularx}
\usepackage{colortbl}
\usepackage{stfloats}
\usepackage{tabularx}
\def\BibTeX{{\rm B\kern-.05em{\sc i\kern-.025em b}\kern-.08em
    T\kern-.1667em\lower.7ex\hbox{E}\kern-.125emX}}
\begin{document}

\title{Unleashing More Actions via Action Compositional Training for VLA Models}

\author{\IEEEauthorblockN{Kai Peng$^{*}$}
\IEEEauthorblockA{\textit{School of Artificial Intelligence} \\
\textit{Shenzhen Technology University} \\
Shenzhen, China\\
2410263047@stumail.sztu.edu.cn}
\and
\IEEEauthorblockN{Jie Lu$^{*}$}
\IEEEauthorblockA{\textit{School of Artificial Intelligence} \\
\textit{Shenzhen Technology University} \\
Shenzhen, China\\
2510263005@stumail.sztu.edu.cn}
\and
\IEEEauthorblockN{Xiaojiang Peng$^{\dagger}$}
\IEEEauthorblockA{\textit{School of Artificial Intelligence} \\
\textit{Shenzhen Technology University} \\
Shenzhen, China\\
pengxiaojiang@sztu.edu.cn}
\thanks{$^{*}$These authors contributed equally to this work.}
\thanks{$^{\dagger}$Corresponding author: Xiaojiang Peng (pengxiaojiang@sztu.edu.cn)}
}

\maketitle

\begin{abstract}
Vision-Language-Action (VLA) models excel at robotic manipulation, driven by the scale and diversity of demonstration data. However, standard training paradigms often cause VLA models to severely overfit to specific behavioral patterns, rendering them unable to generalize to out-of-distribution scenarios even when those scenarios merely require novel combinations of identical sub-skills. While expanding datasets can mitigate this overfitting, acquiring high-quality robot data remains notoriously labor-intensive and cost-prohibitive. To resolve this impasse without expensive human teleoperation and to truly unleash more actions—i.e., enable VLA models to compose known sub-skills into a much broader set of executable behaviors beyond the original demonstrations—we propose ACT-VLA (Action Compositional Training for VLA Models), an offline data augmentation framework that leverages the model’s latent task representations to synthesize novel, physically valid demonstrations directly from existing tasks for policy training. By eliminating additional manual data collection, our method automatically expands the training distribution and mitigates overfitting. We evaluate our approach on challenging manipulation tasks in simulation. Experiments demonstrate that while baseline VLA models generalize poorly due to original distribution overfitting, policies trained with our synthesized data achieve substantially higher success rates, validating that leveraging existing tasks for automated demonstration synthesis provides an effective, scalable, and data-efficient route to broadening VLA generalization.
\end{abstract}

\begin{IEEEkeywords}
Vision-Language-Action models, robotic manipulation,  compositional generalization
\end{IEEEkeywords}

\section{Introduction}
Robotic manipulation has undergone a fundamental transformation with the rise of large-scale imitation learning~\cite{brohan2022rt,o2024open}. By training on extensive collections of human demonstrations, modern robot learning systems have achieved remarkable dexterity across a broad range of manipulation tasks. Among these approaches, Vision-Language-Action (VLA) models~\cite{zitkovich2023rt, kim2024openvla, octo_2023, black2024pi_0, bjorck2025gr00t, pi05} have emerged as a particularly promising paradigm, integrating visual perception, language understanding, and action generation into a unified framework capable of following natural-language instructions and executing complex robotic behaviors.

Despite their impressive performance, the generalization capability of VLA models remains fundamentally constrained by the diversity and coverage of training data. Existing imitation learning pipelines learn manipulation behaviors directly from demonstrations, causing policies to rely heavily on previously observed task distributions. As shown by recent evaluations of VLA robustness and compositional generalization~\cite{wang2024towards,zhou2025libero,Libero-plus}, models often struggle when confronted with novel combinations of familiar skills. Although individual sub-skills may have been successfully learned, the model frequently fails to execute them in unseen sequences because the corresponding demonstrations were absent during training. Since the number of possible task compositions grows combinatorially with the number of available skills, exhaustive demonstration coverage becomes practically impossible.

This challenge is exacerbated by the high cost of robotic data collection~\cite{o2024open}. Unlike vision and language domains, where large-scale datasets can be gathered from readily available internet resources, robotic demonstrations require physical interaction with real systems. Each new task must be collected through teleoperation, involving human supervision, environment preparation, hardware maintenance, and repeated trial execution. As task diversity increases, the cost of acquiring demonstrations grows proportionally, while the space of potential task compositions expands exponentially. Consequently, scaling generalization solely through manual data collection is neither economically nor practically sustainable.

These observations suggest that improving robotic generalization may require not only larger models, but also more scalable approaches to data generation. Recent work has revealed that VLA models encode latent compositional structures that can be exploited to synthesize valid trajectories for previously unseen task combinations~\cite{livlas}. By interpolating internal task representations, such methods are capable of generating novel behaviors without requiring additional demonstrations. This capability presents a promising opportunity: rather than using compositional synthesis solely as an inference-time control mechanism, can it be leveraged as a source of new training data?

Motivated by this question, we propose \textbf{ACT-VLA (Action Compositional Training for VLA Models)}, an offline data augmentation framework that transforms compositional trajectory synthesis into an automatic demonstration generation pipeline. Instead of collecting additional demonstrations from human operators, our method reuses existing task trajectories and synthesizes novel compositional demonstrations that are absent from the original dataset. These generated demonstrations are subsequently incorporated into standard VLA training, enabling the model to learn compositional transitions directly from data.

\begin{figure}[t]
    \centering
    \includegraphics[width=\linewidth]{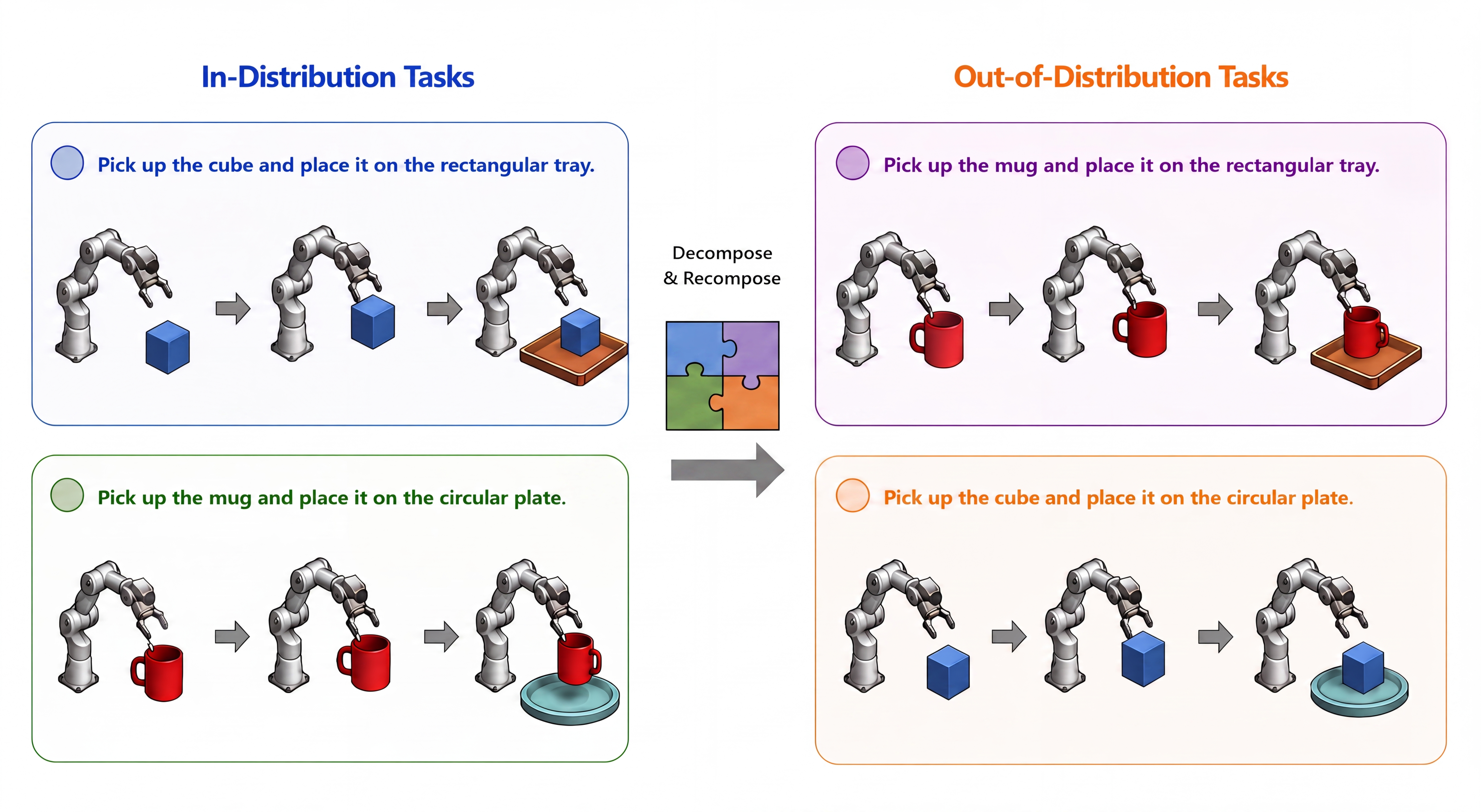}
    \caption{\textbf{Illustration of compositional generalization via 
    object--target re-pairing.} 
    \textbf{Left (In-Distribution):} training tasks contain fixed object--target 
    couplings---\emph{cube}$\to$\emph{rectangular tray} and 
    \emph{mug}$\to$\emph{circular plate}. 
    \textbf{Right (Out-of-Distribution):} by decomposing and recombining the 
    original task structures, we construct novel test configurations never seen 
    during training---\emph{mug}$\to$\emph{rectangular tray} and 
    \emph{cube}$\to$\emph{circular plate} . These re-paired combinations 
    require the model to generalize beyond its memorized associations.}
    \label{fig:p1}
\end{figure}

Fig.~\ref{fig:p1} illustrates the core motivation of this work with a concrete 
example. During standard training, VLA models are exposed to tasks 
with fixed object--target pairings, e.g., always placing the \emph{cube} on the 
\emph{rectangular tray} and the \emph{mug} on the \emph{circular plate}. 
Consequently, the model memorizes these rigid visual--language--action 
associations and fails when confronted with re-paired combinations such as 
placing the \emph{mug} on the \emph{rectangular tray} or the \emph{cube} on the 
\emph{circular plate}. Our method addresses this failure by automatically 
synthesizing training demonstrations for such novel object--target pairings 
directly from existing data, without collecting any new human demonstrations. 
In this way, our method expands the effective coverage of the training 
distribution while requiring no additional teleoperation effort.

We evaluate our method on the LIBERO simulation benchmark~\cite{liu2023libero}. Experimental results show that the synthesized demonstrations significantly improve performance on unseen task combinations compared with baseline VLA training. On out-of-distribution compositional suites, our method achieves absolute gains of $+52.7$ and $+49.0$ percentage points on Spatial-OOD and Goal-OOD, respectively, while maintaining competitive performance on standard benchmarks.

The main contributions of this work are threefold: (i) to the best of our knowledge, we are the first to address the problem of compositional demonstrations for Vision-Language-Action (VLA) models; (ii) we propose ACT-VLA, a plug-and-play data augmentation method that leverages text latent interpolation for compositional demonstration synthesis; and (iii) we achieve state-of-the-art results on the LIBERO-Spatial-OOD and LIBERO-Goal-OOD datasets





\section{RELATED WORK}

\subsection{Vision-Language-Action Models}

Vision-Language-Action (VLA) models integrate visual perception, language understanding, and low-level control into a unified policy architecture. RT-2~\cite{zitkovich2023rt} demonstrated that knowledge acquired from large-scale vision-language pretraining can be transferred to robotic manipulation. Building upon this direction, OpenVLA~\cite{kim2024openvla} provided an open-source framework for large-scale VLA training, while $\pi_0$~\cite{black2024pi_0} introduced a flow-matching formulation for continuous action generation. Octo~\cite{octo_2023} and GR00T N1~\cite{bjorck2025gr00t} further expanded policy scalability through training on increasingly diverse robotic datasets. Beyond scaling model capacity and data volume, recent work has explored orthogonal directions to strengthen VLA capabilities. InstructVLA~\cite{instructvla} introduces vision-language-action instruction tuning that jointly optimizes embodied reasoning and action generation, preserving the flexible reasoning of pretrained VLMs while achieving strong manipulation performance. ReconVLA~\cite{ReconVLA} proposes a reconstructive objective where a diffusion transformer learns to reconstruct manipulated object regions from VLA visual features, implicitly guiding attention toward task-relevant targets for precise control. Spatial Forcing~\cite{spatialforcing} aligns intermediate VLA visual embeddings with geometric representations from pretrained 3D foundation models, implicitly injecting spatial awareness without requiring explicit depth sensors. 3D-VLA~\cite{3d-vla} further integrates a 3D-based LLM with embodied diffusion models to construct a generative world model that links 3D perception, reasoning, and action. These developments have established VLA models as a dominant paradigm for general-purpose robotic manipulation and form the policy foundation considered in this work.

\subsection{Compositional Generalization}
\label{sec:related_compositional}
Compositional generalization, the ability to recombine known behavioral primitives into novel sequences, is a longstanding challenge in robot learning.
Decomposition-based approaches address this by explicitly partitioning demonstrations into reusable skill primitives. DeCo~\cite{chen2026deco} segments manipulation demonstrations into modular atomic tasks based on gripper-object interaction cycles, constructs a reusable skill library, and at inference leverages a vision-language model to parse novel instructions, retrieve relevant skills, and schedule their execution via a spatially-aware chaining module that generates collision-free transitions between consecutive skills. LiLo-VLA~\cite{yang2026lilo} decouples long-horizon execution into a Reaching Module for global transport via motion planning and an object-centric Interaction Module for fine-grained manipulation, enabling zero-shot generalization to novel skill sequences while mitigating cascading failures through closed-loop recovery. Both methods achieve compositional behavior without requiring demonstrations of complete task sequences, but rely on external planners or auxiliary perception components at inference time.

Beyond decomposition-based methods, planning-centric approaches address compositional and long-horizon manipulation from complementary perspectives. Generative Skill Chaining (GSC)~\cite{generative-skill-chaining} learns skill-centric diffusion models and composes their learned distributions to produce long-horizon plans at inference time without requiring explicit task decomposition. Long-VLA~\cite{fan2025long} proposes a phase-aware input masking strategy that adaptively segments subtasks into moving and interaction phases, enabling end-to-end VLA models to handle long-horizon tasks within a unified architecture. Plan-Seq-Learn~\cite{plan-seq-learn} bridges LLM-based high-level planning with RL-based low-level control, solving multi-stage robotic tasks from raw visual input without a predefined skill library. In parallel, search-based planning methods~\cite{search-based} jointly search over parameterized skills using learned skill effect models, while Inner Monologue~\cite{inner-monologue} leverages closed-loop language feedback from LLMs to improve high-level instruction completion. Although effective, these methods typically depend on predefined skill libraries, external planners, or auxiliary reasoning components, whereas our approach internalizes compositional transitions entirely within the VLA's weights through offline data augmentation.

Representation-based approaches instead exploit compositional structures latent within pretrained policies. Li~\cite{livlas} demonstrated that transformer-based VLAs implicitly encode sub-skill semantics within the hidden states of text tokens, and proposed Text Latent Interpolation (TLI) to synthesize smooth behavioral transitions between base skills by steering these internal representations at inference time, without architectural modification or additional training. This result establishes that compositional behaviors can emerge directly from latent representations, an insight that our method builds upon by relocating TLI from the inference stage to an offline data synthesis pipeline.

\subsection{Data-Centric Robot Learning}

Recent robot learning research has explored a variety of data-centric strategies, including dataset construction, supervision relabeling, and training distribution expansion, to improve policy performance and generalization. Large-scale efforts such as RT-1~\cite{brohan2022rt} and Open X-Embodiment~\cite{o2024open} highlight the role of diverse demonstration datasets in training general-purpose manipulation policies.
Beyond dataset scaling, several studies have investigated methods for enriching existing robot data without additional teleoperation. Interleave-VLA~\cite{faninterleave} automatically converts text-only demonstrations into image-text interleaved supervision, providing richer multimodal grounding from existing trajectories. PixelVLA~\cite{liang2026pixelvla} augments demonstrations with automatically generated pixel-level annotations, introducing additional spatial supervision for policy learning. Both approaches improve the utility of existing datasets by generating new forms of supervision without collecting additional demonstrations.
GraspVLA~\cite{graspvla} further demonstrates the effectiveness of billion-scale synthetic action data generated in simulation with photorealistic rendering for pretraining grasping foundation models.
These methods collectively demonstrate that enriching robot training data—through richer supervision~\cite{faninterleave,liang2026pixelvla}, visual augmentation~\cite{scaling-robot-learning}, or large-scale synthetic data~\cite{graspvla}—is an effective paradigm for improving policy capability. Building on this data-centric philosophy, our method leverages existing demonstrations to synthesize novel \emph{compositional} task trajectories by recombining behavioral segments across different tasks, thereby expanding the set of trainable task combinations without additional data collection.

\section{Method}

\begin{figure*}[t]
    \centering
    \includegraphics[width=\textwidth]{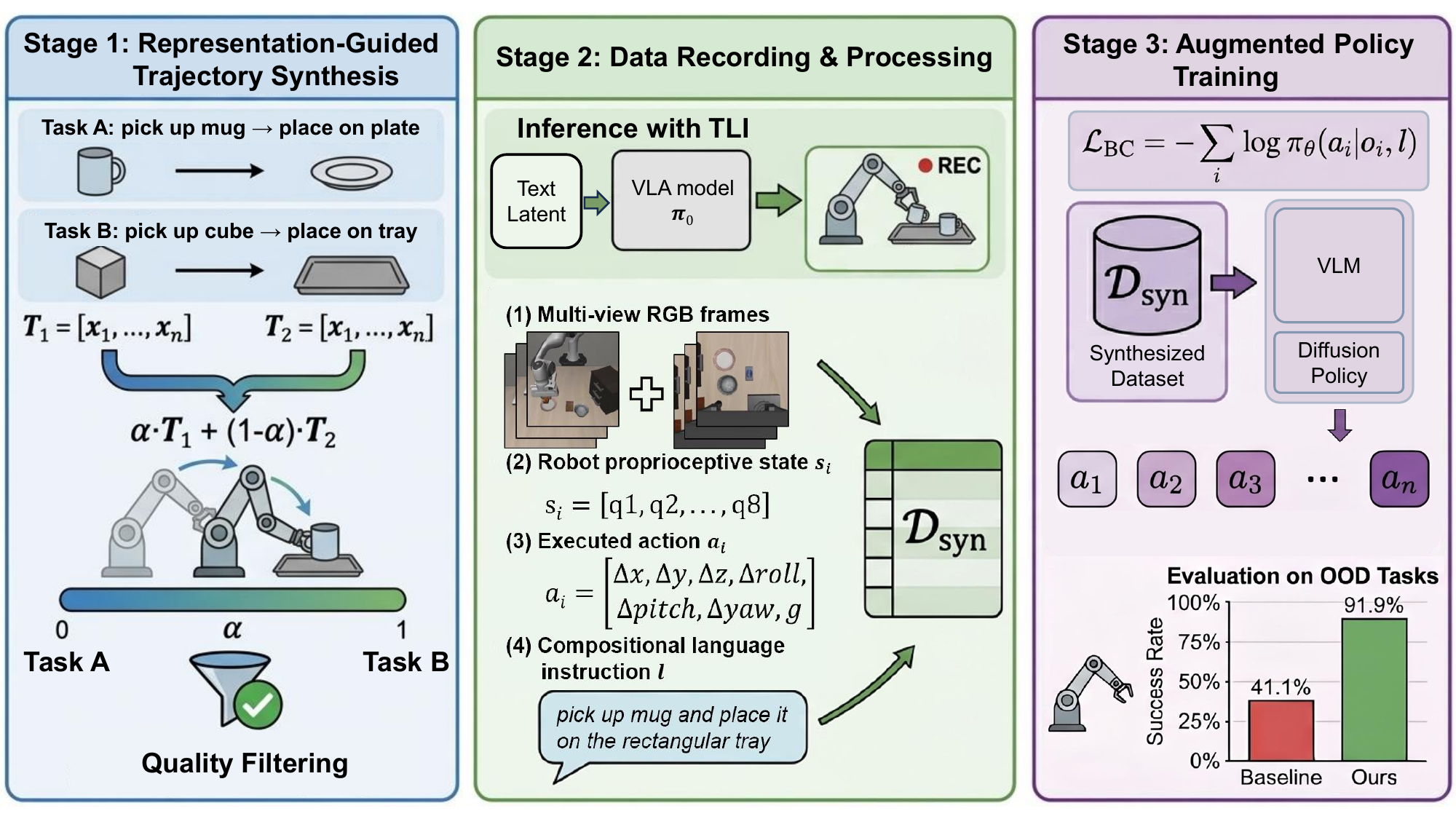}
    \caption{\textbf{The overall pipeline of our 
    proposed method.} Our approach consists of three 
    stages: (1) representation-guided trajectory 
    synthesis, (2) data recording and processing, 
    and (3) training on the augmented dataset. 
    Synthesized demonstrations are filtered by the 
    physical simulator before being incorporated 
    into training.}
    \label{fig:pipeline}
\end{figure*}

\subsection{Overview}
Our method transforms compositional trajectory 
synthesis from an inference-time mechanism into an 
offline data generation pipeline. The core idea is to 
exploit the compositional structures already encoded 
in a pretrained VLA to produce training demonstrations 
for task combinations absent from the original 
dataset, without requiring any additional human 
teleoperation. Our method consists of three stages: 
we first apply representation-guided trajectory 
synthesis to generate demonstrations of compositional 
tasks. The successful rollouts are recorded and processed into a structured training format. Finally, 
we train VLA models on the augmented dataset. 
The overall pipeline is illustrated in 
Fig.~\ref{fig:pipeline}.

\subsection{Compositional Demonstration Synthesis}

\subsubsection{Synthesis via Latent Interpolation}
To generate demonstrations for novel compositional tasks 
without human teleoperation, we leverage the \textbf{text latent 
interpolation (TLI)} mechanism introduced by~\cite{livlas}. 
Transformer-based VLAs encode task semantics within the 
hidden states of text tokens. By averaging these states 
across demonstrated episodes, one can extract a 
task-specific \textit{text latent} $\mathcal{T}$ that 
encapsulates the required behavioral context. 

To synthesize a continuous transition between two base tasks, we interpolate their respective text latents, $\mathcal{T}^1$ and $\mathcal{T}^2$, to steer the model's internal representations. Specifically, at each timestep $i$, the text hidden states are modified as follows:
\begin{equation}
h^T(i) \leftarrow h^T(i) + (1-\alpha)\mathcal{T}^1 + \alpha\mathcal{T}^2 - \left[ (1-\alpha)\mathcal{T}^2 + \alpha\mathcal{T}^1 \right]
\end{equation}
where the transition speed $\alpha = i/\lambda$ is linearly scaled and clipped to $[0, 1]$. At the start of the episode ($\alpha \approx 0$), Task~2's context is suppressed and subtracted from the residual stream, which reinforces Task~1's behavior. As the episode progresses ($\alpha \to 1$), Task~2's context is gradually injected into the residual stream while Task~1's context fades out, yielding a physically continuous compositional trajectory.

\subsubsection{From Inference-Time Intervention to Training-Time Data Synthesis}
While the interpolation mechanism is adopted from 
prior work~\cite{livlas}, the core contribution of our 
method lies not in the interpolation operation itself, 
but in \textit{repurposing} it as a systematic data 
generation engine and addressing the non-trivial 
challenges that arise when synthesized trajectories 
are used for policy training. We highlight three key 
design dimensions that distinguish our approach from a 
simple ``move TLI to training'':

\begin{itemize}
    \item \textbf{Instruction Design.} 
    Inference-time TLI requires no explicit task 
    instruction, as the latent steering signal alone 
    guides behavior. In contrast, training-time 
    synthesis demands a language instruction $\ell$ 
    that faithfully describes the compositional task 
    across all timesteps. We derive $\ell$ directly 
    from the base task instructions by splitting and 
    recombining their textual components to reflect 
    the new task structure (see 
    Section~\ref{sec:data_recording}).
    
    \item \textbf{Quality Control.} 
    When TLI is applied at inference, failed rollouts 
    are discarded and the episode is simply retried. 
    For training data, however, including even a small 
    fraction of failed trajectories can propagate 
    incorrect state-action associations and degrade 
    the policy. We therefore introduce a dual 
    filtering strategy combining task completion 
    verification with a step-budget constraint to 
    ensure only high-quality, physically valid 
    trajectories enter the training set (see 
    Section~\ref{sec:filtering}).
    
    \item \textbf{Training Dynamics.} 
    Simply adding synthesized data to the training set 
    does not guarantee improved generalization---it 
    introduces a distribution shift that must be 
    carefully managed. We maintain parity between 
    original and synthesized data volumes at the suite 
    level to prevent the synthetic distribution from 
    overwhelming the original task distribution and 
    causing catastrophic forgetting (see 
    Section~\ref{sec:synthesis_scale}).
\end{itemize}

These design decisions collectively transform TLI from 
a per-episode inference hack into a principled data 
augmentation framework. The empirical results in 
Section~\ref{sec:experiments} confirm that this 
training-time approach not only matches but 
substantially exceeds the OOD performance of 
inference-time TLI, while eliminating all deployment-time 
computational overhead.

\subsubsection{Trajectory Quality Filtering}
\label{sec:filtering}
Not all synthesized rollouts are suitable for 
training. Since TLI operates by steering latent 
representations rather than guaranteeing physically 
correct behavior, some rollouts may fail to complete 
one or both sub-tasks. We treat the physical simulator 
as a rigorous filter, retaining only those rollouts in 
which the policy successfully completes the full 
compositional task according to the strict task 
completion criteria defined in LIBERO. Rollouts that 
fail at any stage are discarded. This filtering step 
is essential: naively including failed rollouts would 
introduce incorrect state-action associations into the 
training set, potentially degrading the policy's base 
skill retention.

\subsubsection{Synthesis Scale}
\label{sec:synthesis_scale}
Following the structure of the original LIBERO task 
suites, we synthesize demonstrations for each 
compositional task such that the resulting dataset 
matches the scale of the corresponding original suite. 
Specifically, each suite contains 10 tasks, and for 
each task we collect approximately 30--40 successful 
trajectories, yielding a total of roughly 300--400 
synthesized demonstrations per OOD suite. To ensure 
trajectory quality, we further restrict the maximum 
episode length during evaluation-based recording: only 
rollouts that complete the compositional task within a 
predefined step budget are retained. This step-based 
filtering complements the task completion check and 
helps exclude trajectories with unnecessarily long or 
redundant segments, which could dilute the training 
signal. The final synthesized dataset 
$\mathcal{D}_\text{syn}$ is then different with the 
original demonstrations, as described in Section~\ref{sec:augmented_training}.

\subsection{Data Recording and Processing}
\label{sec:data_recording}
We implement a data recording module to convert 
successful synthesized rollouts into a structured 
training format. At each timestep $i$, the module 
records the RGB observations $o_i$ from all available 
camera views, the robot proprioceptive state $s_i$ 
(comprising joint positions and end-effector pose), 
the executed action $a_i$, and the natural language 
instruction $\ell$ for the compositional task.

A key design decision concerns the formulation of 
$\ell$. Since the compositional trajectory is 
synthesized by combining sub-skill segments from two 
base tasks, we derive the corresponding instruction 
by splitting and recombining the textual components 
of the original task instructions to reflect the new 
task structure (e.g., re-pairing the object from 
Task~A with the target from Task~B). This ensures 
that the language supervision remains semantically 
consistent with the synthesized trajectory 
throughout all timesteps. The recorded 
data is strictly formatted to match the demonstration 
structure of the original dataset, requiring no 
modifications to the training pipeline.

\subsection{Augmented Policy Training}
\label{sec:augmented_training}


We train the VLA model on $\mathcal{D}_\text{syn}$ using standard 
behavioral cloning with the same loss function and 
hyperparameters as the baseline VLA model, with no 
modifications to the architecture. This design ensures 
that compositional capability is acquired entirely 
through data exposure rather than architectural 
intervention. Since compositional transitions are 
internalized into model weights during training, no 
inference-time latent manipulation or external 
planners are required at deployment.

\section{Experiments}
\label{sec:experiments}

\subsection{Experimental Setup}

\subsubsection{Benchmark}
\begin{figure}[t]
    \centering
    \includegraphics[width=\linewidth]{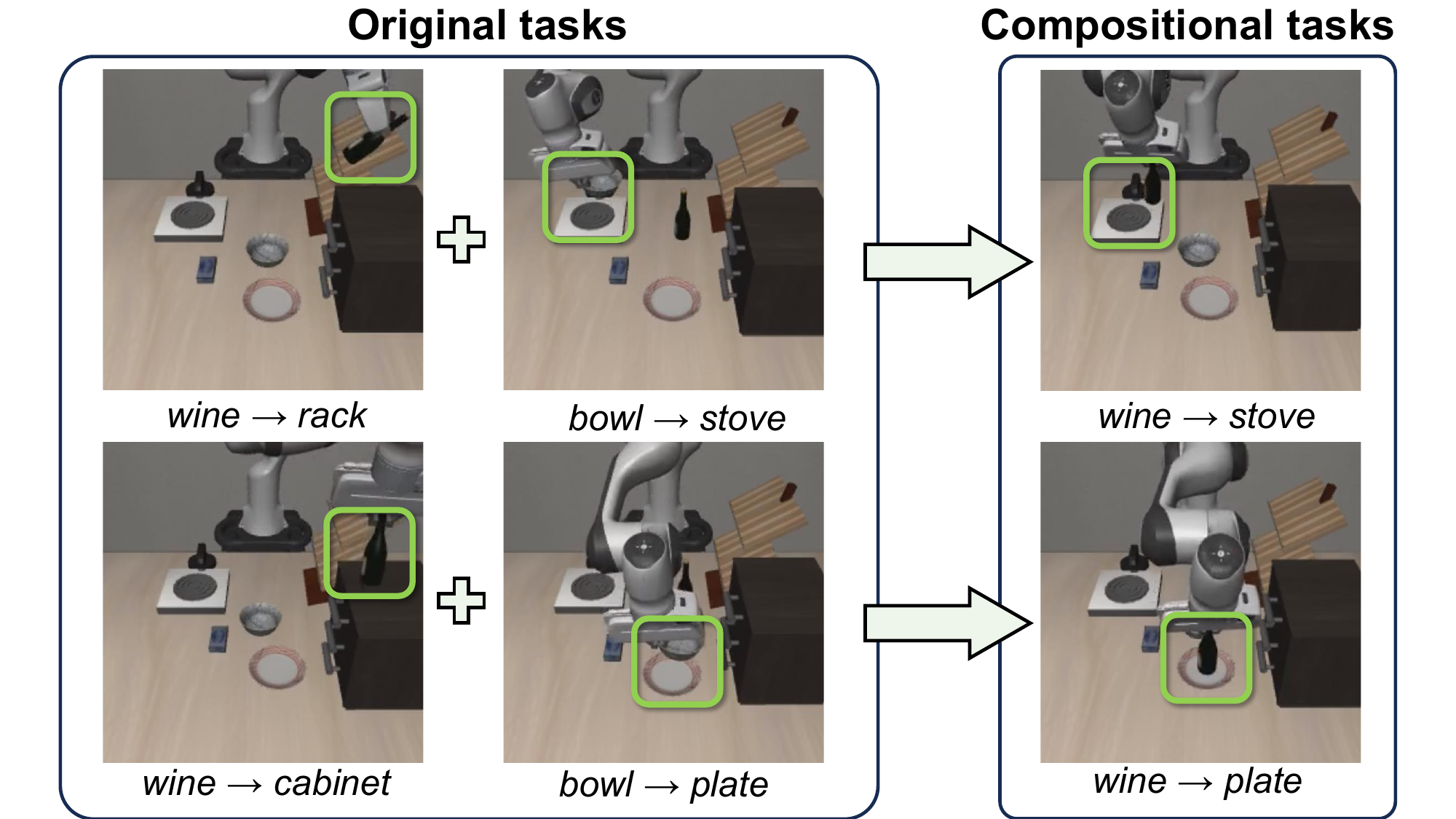}
    \caption{\textbf{Illustration of compositional task construction.} 
Each compositional task (right) is formed by combining the 
object from one original task with the target location from 
another, following the task construction protocol of~\cite{livlas}. 
These re-paired combinations are absent from the training set, 
serving as out-of-distribution evaluation scenarios.}
    \label{fig:exp}
\end{figure}

We evaluate our method on the LIBERO 
benchmark~\cite{liu2023libero}, a comprehensive 
simulation suite designed to assess knowledge 
transfer and generalization in robot manipulation. 
LIBERO consists of four task suites: 
LIBERO-Spatial, LIBERO-Object, LIBERO-Goal, and 
LIBERO-10, each comprising 10 tasks that test 
different aspects of manipulation knowledge. To 
evaluate compositional generalization, we 
additionally construct two out-of-distribution 
evaluation suites, LIBERO-Spatial-OOD and 
LIBERO-Goal-OOD, containing novel task compositions 
that are absent from the original training 
demonstrations but decomposable into sub-skills 
present in the training set, as illustrated in Fig.~\ref{fig:exp}.
\subsubsection{Baselines}
We compare our method against two categories of 
baselines. For state-of-the-art comparison, we 
include UniVLA~\cite{bu2025univla}, 
OpenVLA-OFT~\cite{kim2025fine}, 
GR00T-N1~\cite{bjorck2025gr00t}, and 
$\pi_{0.5}$~\cite{pi05}, all trained on the 
original LIBERO demonstrations under the same 
protocol. We additionally include $\pi_0$ with 
text latent interpolation applied at inference 
time~\cite{livlas} as a direct comparison against 
the inference-time alternative to our training-time 
synthesis approach.

\subsubsection{Evaluation Protocol}
Each task is evaluated over 3 random seeds, each 
with 50 episodes (150 episodes total per task). 
We report the mean success rate (\%) over all 
tasks within each suite. All models are trained 
for 30k steps under identical settings.

\subsection{Main Results}

\begin{table*}[tb]
\centering
\caption{Performance comparison on LIBERO 
benchmark, including both in-distribution and 
out-of-distribution suites. Results are success 
rates (\%) averaged across tasks. $\dagger$ denotes 
inference-time latent manipulation without policy 
retraining. \textbf{Bold} indicates best OOD 
performance. $\Delta$ is computed relative to 
$\pi_{0.5}$ trained on original data. AVG is the 
unweighted mean of all six suite-level scores.}
\label{tab:main}
\begin{tabularx}{0.8\textwidth}{X cccc | cc | c}
\toprule
\textbf{Model} & 
\textbf{Spatial} & \textbf{Goal} & 
\textbf{Object} & \textbf{Long} & 
\textbf{Spatial OOD} & \textbf{Goal OOD} & 
\textbf{AVG} \\
\midrule
UniVLA~\cite{bu2025univla}       
& 96.5 & 95.6 & 96.8 & 92.0 
& 11.0 & 32.0 & 70.7 \\
OpenVLA-OFT~\cite{kim2025fine}  
& 97.6 & 97.9 & 98.4 & 94.5 
& 0.0  & 1.0  & 64.9 \\
GR00T-N1~\cite{bjorck2025gr00t}  
& 94.1 & 98.3 & 99.1 & 93.2 
& 17.9 & 24.1 & 71.1 \\
$\pi_{0.5}$~\cite{pi05}          
& 98.8 & 98.5 & 99.3 & 92.9 
& 35.5 & 46.6 & 78.6 \\
$\pi_0^{\dagger}$~\cite{livlas}  
& 96.8 & 98.8 & 95.8 & 85.2 
& 81.0 & 85.0 & 90.4 \\
\midrule
\rowcolor{gray!20}
\textbf{Ours} ($\pi_{0.5}$)      
& 98.8 & 98.5 & 99.3 & 92.9 & 
\textbf{88.2}{\scriptsize~($+52.7$)} & 
\textbf{95.6}{\scriptsize~($+49.0$)} & 
\textbf{95.6}{\scriptsize~($+17.0$)} \\
\bottomrule
\end{tabularx}
\end{table*}

Our method delivers substantial improvements on the compositional out-of-distribution suites (as shown in Table~\ref{tab:main}). Compared to $\pi_{0.5}$ trained on original data only, performance on LIBERO-Spatial-OOD increases from 35.5\% to 88.2\% and on LIBERO-Goal-OOD from 46.6\% to 95.6\%, corresponding to absolute gains of 52.7 and 49.0 percentage points, respectively. These results validate that generating compositional demonstrations from existing datasets is an effective route to broadening VLA generalization to novel task compositions.

The AVG column reports the unweighted mean of all six suite-level success rates, treating each suite equally regardless of its internal task count. We choose unweighted averaging to prevent suites with more tasks from dominating the aggregate metric, thereby ensuring that improvements on OOD suites (which contain fewer tasks than the standard suites) are not numerically diluted.

Compared to $\pi_0$ with TLI applied at inference time, our approach achieves higher OOD success rates (88.2\% vs. 81.0\% on Spatial-OOD, 95.6\% vs. 85.0\% on Goal-OOD) while requiring no deployment-time intervention, demonstrating that training-time synthesis is a more effective 
strategy than inference-time latent steering. Averaging across all six evaluation suites, our 
method achieves an AVG score of 95.6\%, outperforming all baselines including UniVLA (70.7\%), OpenVLA-OFT (64.9\%), GR00T-N1 (71.1\%), $\pi_{0.5}$ without synthesis (78.6\%), and $\pi_0$ with inference-time TLI (90.4\%).

\subsection{Ablation Study}

To isolate the contribution of synthesized data 
and examine the difference between training-time 
and inference-time compositional synthesis, 
Table~\ref{tab:ablation} presents controlled 
comparisons conducted on the $\pi_0$ backbone, 
which allows a fair same-architecture evaluation 
independent of backbone capacity differences.

Without any compositional demonstrations, $\pi_0$ 
trained on original data achieves only 2.0\% and 
16.0\% on Spatial-OOD and Goal-OOD respectively, 
confirming that standard imitation learning fails 
to generalize to unseen task compositions even 
when all constituent sub-skills have been 
individually mastered. Incorporating our 
synthesized data substantially improves these 
scores to 85.9\% and 90.5\%, demonstrating that 
synthesized compositional demonstrations provide 
an effective training signal independent of the 
specific VLA backbone.

\begin{table}[h]
\centering
\caption{Ablation study on $\pi_0$ backbone. 
$\dagger$ denotes inference-time method reproduced 
from Table~\ref{tab:main} for direct comparison. 
$\Delta$ is computed relative to $\pi_0$ trained 
on original data only.}
\label{tab:ablation}
\setlength{\tabcolsep}{3.5pt}
\resizebox{\linewidth}{!}{\begin{tabular}{l l cc cc}
\toprule
\multirow{2}{*}{\textbf{Model}} & 
\multirow{2}{*}{\textbf{Data}} & 
\multicolumn{2}{c}{\textbf{In-Dist.}} & 
\multicolumn{2}{c}{\textbf{OOD}} \\
\cmidrule(lr){3-4} \cmidrule(lr){5-6}
& & \textit{Spatial} & \textit{Goal} & 
\textit{Spatial} & \textit{Goal} \\
\midrule
$\pi_0$ & Original only    
& 96.8 & 98.8 & 2.0  & 16.0 \\
$\pi_0^{\dagger}$ & TLI at inference    
& 96.8 & 98.8 & 81.0 & 85.0 \\
\rowcolor{gray!20}$\pi_0$ & Orig. + Synth.    
& 95.6 & 96.1 & 
85.9{\scriptsize~($\Delta+83.9$)} & 
90.5{\scriptsize~($\Delta+74.5$)} \\
\bottomrule
\end{tabular}}
\end{table}
A direct comparison between $\pi_0$ with TLI at 
inference time (81.0\%, 85.0\%) and $\pi_0$ 
trained with our synthesized data (85.9\%, 90.5\%) 
holds the backbone constant, isolating the effect 
of when compositional synthesis is applied. 
Training-time synthesis achieves higher OOD 
performance while eliminating the per-step 
computational overhead of inference-time latent 
manipulation, confirming that internalizing 
compositional transitions into model weights 
through offline data augmentation is both more 
effective and more practical for deployment.

\section{Conclusion}

We presented Action Compositional Training, an offline data augmentation 
framework that repurposes text latent interpolation as an automated 
demonstration synthesis engine. By addressing the challenges of instruction 
design, quality control, and training dynamics, our method enables VLA models 
to internalize compositional skill transitions directly into their weights 
without inference-time intervention. Experiments on LIBERO demonstrate 
substantial OOD improvements while maintaining competitive in-distribution 
performance. Our 
method requires no additional teleoperation, architectural changes, or 
external planners, offering a practical route toward broadening VLA 
compositional generalization.

\section{Limitations and Future Work}
While Action Compositional Training provides a data-efficient pipeline for VLA generalization, several limitations remain.
First, our validation focuses on sequential two-skill compositions within tabletop constraints. Although text latent interpolation is theoretically extensible to longer chains, recursively organizing and filtering multi-skill rollouts introduces exponential optimization challenges.
Second, synthesis quality depends on the semantic disentanglement of the pretrained VLA backbone. Deploying our method on weaker base models might bottleneck performance, suggesting that integrating representation learning could be beneficial.
Finally, all experiments were conducted in simulation. Although our dual-filtering strategy ensures physical validity to mitigate distribution shifts, the sim-to-real gap on physical hardware warrants further evaluation in future work.

\bibliographystyle{IEEEtran}   
\bibliography{ref}         

@inproceedings{zitkovich2023rt,
  title={Rt-2: Vision-language-action models transfer web knowledge to robotic control},
  author={Zitkovich, Brianna and Yu, Tianhe and Xu, Sichun and Xu, Peng and Xiao, Ted and Xia, Fei and Wu, Jialin and Wohlhart, Paul and Welker, Stefan and Wahid, Ayzaan and others},
  booktitle={Conference on Robot Learning (CoRL)},
  pages={2165--2183},
  year={2023},
  organization={PMLR}
}

@inproceedings{kim2024openvla,
  title={Openvla: An open-source vision-language-action model},
  author={Kim, Moo Jin and Pertsch, Karl and Karamcheti, Siddharth and Xiao, Ted and Balakrishna, Ashwin and Nair, Suraj and Rafailov, Rafael and Foster, Ethan and Lam, Grace and Sanketi, Pannag and others},
  booktitle={Conference on Robot Learning (CoRL)},
  pages={2679--2713},
  year={2025},
  publisher={PMLR},
}

@article{black2024pi_0,
  title={{$\pi_0$}: A Vision-Language-Action Flow Model 
  for General Robot Control},
  author={Black, Kevin and Brown, Noah and Driess, Danny 
  and Esmail, Adnan and Equi, Michael and Finn, Chelsea 
  and Fusai, Niccolo and Groom, Lachy and Hausman, Karol 
  and Ichter, Brian and others},
  journal={arXiv preprint arXiv:2410.24164},
  year={2024}
}

@inproceedings{livlas,
  author    = {Quanyi Li},
  title     = {VLAs are Confined yet Capable of Generalizing to Novel Instructions},
  booktitle = {Proceedings of the International Joint Conference on Artificial Intelligence (IJCAI)},
  year      = {2026},
}

@article{Libero-plus,
  title={Libero-plus: In-depth robustness analysis of vision-language-action models},
  author={Fei, Senyu and Wang, Siyin and Shi, Junhao and Dai, Zihao and Cai, Jikun and Qian, Pengfang and Ji, Li and He, Xinzhe and Zhang, Shiduo and Fei, Zhaoye and others},
  journal={arXiv preprint arXiv:2510.13626},
  year={2025}
}

@article{zhou2025libero,
  title={LIBERO-PRO: Towards Robust and Fair Evaluation of Vision-Language-Action Models Beyond Memorization},
  author={Zhou, Xueyang and Xu, Yangming and Tie, Guiyao and Chen, Yongchao and Zhang, Guowen and Chu, Duanfeng and Zhou, Pan and Sun, Lichao},
  journal={arXiv preprint arXiv:2510.03827},
  year={2025}
}

@ARTICLE{chen2026deco,
  author={Chen, Zixuan and Yin, Junhui and Chen, Yangtao and Huo, Jing and Tian, Pinzhuo and Shi, Jieqi and Hou, Yiwen and Li, Yinchuan and Gao, Yang},
  journal={IEEE Robotics and Automation Letters}, 
  title={DeCo: Task Decomposition and Skill Composition for Zero-Shot Generalization in Long-Horizon 3D Manipulation}, 
  year={2026},
  pages={5049-5056},
  doi={10.1109/LRA.2026.3666363}}

@article{liu2023libero,
  title={Libero: Benchmarking knowledge transfer for lifelong robot learning},
  author={Liu, Bo and Zhu, Yifeng and Gao, Chongkai and Feng, Yihao and Liu, Qiang and Zhu, Yuke and Stone, Peter},
  journal={Advances in Neural Information Processing Systems},
  volume={36},
  pages={44776--44791},
  year={2023}
}

@inproceedings{octo_2023,
    title={Octo: An Open-Source Generalist Robot Policy},
    author = {{Octo Model Team} and Dibya Ghosh and Homer Walke and Karl Pertsch and Kevin Black and Oier Mees and Sudeep Dasari and Joey Hejna and Charles Xu and Jianlan Luo and Tobias Kreiman and {You Liang} Tan and Lawrence Yunliang Chen and Pannag Sanketi and Quan Vuong and Ted Xiao and Dorsa Sadigh and Chelsea Finn and Sergey Levine},
    booktitle = {Proceedings of Robotics: Science and Systems},
    address  = {Delft, Netherlands},
    year = {2024},
}

@article{bjorck2025gr00t,
  title={Gr00t n1: An open foundation model for generalist humanoid robots},
  author={Bjorck, Johan and Casta{\~n}eda, Fernando and Cherniadev, Nikita and Da, Xingye and Ding, Runyu and Fan, Linxi and Fang, Yu and Fox, Dieter and Hu, Fengyuan and Huang, Spencer and others},
  journal={arXiv preprint arXiv:2503.14734},
  year={2025}
}

@article{wang2024towards,
  title={Towards testing and evaluating vision-language-action models for robotic manipulation: An empirical study},
  author={Wang, Zhijie and Zhou, Zhehua and Song, Jiayang and Huang, Yuheng and Shu, Zhan and Ma, Lei},
  journal={arXiv preprint arXiv:2409.12894},
  volume={1},
  year={2024}
}

@article{yang2026lilo,
  title={LiLo-VLA: Compositional long-horizon manipulation via linked object-centric policies},
  author={Yang, Yue and Cheng, Shuo and Fang, Yu and Bharadhwaj, Homanga and Ding, Mingyu and Bertasius, Gedas and Szafir, Daniel},
  journal={arXiv preprint arXiv:2602.21531},
  year={2026}
}

@article{brohan2022rt,
  title={Rt-1: Robotics transformer for real-world control at scale},
  author={Brohan, Anthony and Brown, Noah and Carbajal, Justice and Chebotar, Yevgen and Dabis, Joseph and Finn, Chelsea and Gopalakrishnan, Keerthana and Hausman, Karol and Herzog, Alex and Hsu, Jasmine and others},
  journal={arXiv preprint arXiv:2212.06817},
  year={2022}
}

@inproceedings{o2024open,
  title={Open x-embodiment: Robotic learning datasets and rt-x models: Open x-embodiment collaboration 0},
  author={O’Neill, Abby and Rehman, Abdul and Maddukuri, Abhiram and Gupta, Abhishek and Padalkar, Abhishek and Lee, Abraham and Pooley, Acorn and Gupta, Agrim and Mandlekar, Ajay and Jain, Ajinkya and others},
  booktitle={2024 IEEE International Conference on Robotics and Automation (ICRA)},
  pages={6892--6903},
  year={2024},
  organization={IEEE}
}

@inproceedings{faninterleave,
  title={Interleave-VLA: Enhancing Robot Manipulation with Image-Text Interleaved Instructions},
  author={Fan, Cunxin and Jia, Xiaosong and Sun, Yihang and Wang, Yixiao and Wei, Jianglan and Gong, Ziyang and Zhao, Xiangyu and Tomizuka, Masayoshi and Yang, Xue and Yan, Junchi and others},
  booktitle={International Conference on Learning Representations},
  year={2026}
}

@inproceedings{liang2026pixelvla,
  title={PixelVLA: Advancing Pixel-level Understanding in Vision-Language-Action Model},
  author={Liang, Wenqi and Sun, Gan and He, Yao and Dong, Jiahua and Dai, Suyan and Laptev, Ivan and Khan, Salman and Cong, Yang},
  booktitle={International Conference on Learning Representations},
  year={2026}
}

@article{bu2025univla,
  title={Univla: Learning to act anywhere with task-centric latent actions},
  author={Bu, Qingwen and Yang, Yanting and Cai, Jisong and Gao, Shenyuan and Ren, Guanghui and Yao, Maoqing and Luo, Ping and Li, Hongyang},
  journal={arXiv preprint arXiv:2505.06111},
  year={2025}
}

@inproceedings{pi05,
  title={{{$\pi_{0.5}$}: A Vision-Language-Action Model with Open-World Generalization}},
  author={Black, Kevin and Brown, Noah Reland and Darpinian, James and Dhabalia, Karan and Driess, Danny and Esmail, Adnan and Equi, Michael Robert and Finn, Chelsea and Fusai, Niccolo and Galliker, Manuel Y and others},
  booktitle={Conference on Robot Learning (CoRL)},
  year={2025}
}

@article{kim2025fine,
  title={Fine-Tuning Vision-Language-Action Models: Optimizing Speed and Success},
  author={Kim, Moo Jin and Finn, Chelsea and Liang, Percy},
  journal={arXiv preprint arXiv:2502.19645},
  year={2025}
}

@inproceedings{instructvla,
  title={{InstructVLA}: Vision-Language-Action Instruction Tuning from Understanding to Manipulation},
  author={Yang, Shuai and Li, Hao and Chen, Yilun and Wang, Bin and Tian, Yang and Wang, Tai and Wang, Hanqing and Zhao, Feng and Liao, Yiyi and Pang, Jiangmiao},
  booktitle={International Conference on Learning Representations (ICLR)},
  year={2026}
}

@inproceedings{ReconVLA,
  title={ReconVLA: Reconstructive Vision-Language-Action Model as Effective Robot Perceiver},
  author={Wenxuan Song and Ziyang Zhou and Han Zhao and Jiayi Chen and Pengxiang Ding and Haodong Yan and Yuxin Huang and Feilong Tang and Donglin Wang and Haoang Li},
  booktitle={AAAI Conference on Artificial Intelligence},
  year={2025},
}

@article{spatialforcing,
  title={Spatial forcing: Implicit spatial representation alignment for vision-language-action model},
  author={Li, Fuhao and Song, Wenxuan and Zhao, Han and Wang, Jingbo and Ding, Pengxiang and Wang, Donglin and Zeng, Long and Li, Haoang},
  journal={arXiv preprint arXiv:2510.12276},
  year={2025}
}

@article{3d-vla,
  title={3d-vla: A 3d vision-language-action generative world model},
  author={Zhen, Haoyu and Qiu, Xiaowen and Chen, Peihao and Yang, Jincheng and Yan, Xin and Du, Yilun and Hong, Yining and Gan, Chuang},
  journal={arXiv preprint arXiv:2403.09631},
  year={2024}
}

@article{graspvla,
  title={Graspvla: a grasping foundation model pre-trained on billion-scale synthetic action data},
  author={Deng, Shengliang and Yan, Mi and Wei, Songlin and Ma, Haixin and Yang, Yuxin and Chen, Jiayi and Zhang, Zhiqi and Yang, Taoyu and Zhang, Xuheng and Zhang, Wenhao and others},
  journal={arXiv preprint arXiv:2505.03233},
  year={2025}
}

@inproceedings{generative-skill-chaining,
  title={Generative Skill Chaining: Long-Horizon Skill Planning with Diffusion Models},
  author={Utkarsh Aashu Mishra and Shangjie Xue and Yongxin Chen and Danfei Xu},
  booktitle={Conference on Robot Learning},
  year={2023},
}

@article{fan2025long,
  title={Long-vla: Unleashing long-horizon capability of vision language action model for robot manipulation},
  author={Fan, Yiguo and Ding, Pengxiang and Bai, Shuanghao and Tong, Xinyang and Zhu, Yuyang and Lu, Hongchao and Dai, Fengqi and Zhao, Wei and Liu, Yang and Huang, Siteng and others},
  journal={arXiv preprint arXiv:2508.19958},
  year={2025}
}

@article{plan-seq-learn,
  title={Plan-seq-learn: Language model guided rl for solving long horizon robotics tasks},
  author={Dalal, Murtaza and Chiruvolu, Tarun and Chaplot, Devendra and Salakhutdinov, Ruslan},
  journal={arXiv preprint arXiv:2405.01534},
  year={2024}
}

@INPROCEEDINGS{search-based,
  author={Liang, Jacky and Sharma, Mohit and LaGrassa, Alex and Vats, Shivam and Saxena, Saumya and Kroemer, Oliver},
  booktitle={2022 International Conference on Robotics and Automation (ICRA)}, 
  title={Search-Based Task Planning with Learned Skill Effect Models for Lifelong Robotic Manipulation}, 
  year={2022},
  pages={6351-6357},
  doi={10.1109/ICRA46639.2022.9811575}}

@article{scaling-robot-learning,
  title={Scaling robot learning with semantically imagined experience},
  author={Yu, Tianhe and Xiao, Ted and Stone, Austin and Tompson, Jonathan and Brohan, Anthony and Wang, Su and Singh, Jaspiar and Tan, Clayton and Peralta, Jodilyn and Ichter, Brian and others},
  journal={arXiv preprint arXiv:2302.11550},
  year={2023}
}

@article{inner-monologue,
  title={Inner monologue: Embodied reasoning through planning with language models},
  author={Huang, Wenlong and Xia, Fei and Xiao, Ted and Chan, Harris and Liang, Jacky and Florence, Pete and Zeng, Andy and Tompson, Jonathan and Mordatch, Igor and Chebotar, Yevgen and others},
  journal={arXiv preprint arXiv:2207.05608},
  year={2022}
}
\end{document}